\documentclass[journal]{IEEEtran}
\usepackage{amsmath,amsfonts}
\usepackage{algorithmic}
\usepackage{algorithm}
\usepackage{array}
\usepackage[caption=false,font=normalsize,labelfont=sf,textfont=sf]{subfig}
\usepackage{textcomp}
\usepackage{stfloats}
\usepackage{url}
\usepackage{verbatim}
\usepackage{graphicx}
\usepackage{cite}
\usepackage[nolist,nohyperlinks]{acronym}
\usepackage{comment}
\usepackage{adjustbox}
\usepackage{multirow, multicol}
\usepackage{orcidlink}
\usepackage{balance}
\usepackage{clipboard}
\newclipboard{myclipboard}

\begin{acronym}
\acro{CNN}{Convolutional Neural Network}
\acro{DNN}{Deep Neural Network}
\acro{FR}{Full-Reference}
\acro{FPS}{Farthest Point Sampling}
\acro{GAP}{Global Average Pooling}
\acro{GCN}{Graph Convolutional Network}
\acro{GMP}{Global Max Pooling}
\acro{G-PCC}{Geometry-based Point Cloud Compression}
\acro{GVP}{Global Variance Pooling}
\acro{HVS}{Human Vision System}
\acro{k-NN}{k-Nearest Neighbour}
\acro{KRCC}{Kendall Rank Correlation Coefficient}
\acro{LiDAR}{Light Detection and Ranging}
\acro{ML}{Lachine Learning}
\acro{MLP}{multilayer perceptron}
\acro{MOS}{mean opinion score}
\acro{MPED}{Multiscale Potential Energy Discrepancy}
\acro{MSE}{Mean Squared Error}
\acro{NR}{No-Reference}
\acro{NSS}{Natural Scene Statistics}
\acro{PCQA}{Point Cloud Quality Assessment}
\acro{PLCC}{Pearson Linear Correlation Coefficient}
\acro{PST}{Patch-based Structure and Texture}
\acro{QoE}{Quality of Experience}
\acro{RMSE}{Root Mean Squared Error}
\acro{RF}{Random Forest}
\acro{RR}{Reduced-Reference}
\acro{SFE}{Structure Feature Extractor}
\acro{TFE}{Texture Feature Extractor}
\acro{SRCC}{Spearman Rank Order Correlation Coefficient}
\acro{SVR}{Support Vector Regression}
\acro{VGG}{Visual Geometry Group}
\acro{V-PCC}{Video-based Point Cloud Compression}
\acro{VR}{Virtual Reality}
\end{acronym}

\definecolor{red_cool}{rgb}{0.5, 0.0, 0.0}

\newcommand{\modified}[1]{{\textbf{\textcolor{red_cool}{#1}}}}

\begin{document}
\title{Low-Complexity Patch-based No-Reference Point Cloud Quality Metric exploiting Weighted Structure and Texture Features}

\author{Michael~Neri~\orcidlink{0000-0002-6212-9139},~\IEEEmembership{Member~IEEE}, Federica~Battisti,~\IEEEmembership{Senior Member, IEEE}~\orcidlink{0000-0002-0846-5879} 
\thanks{
M. Neri is with the with the Faculty of Information Technology and Communication Sciences, Tampere University, Korkeakoulunkatu 1, 33720, Tampere, Finland. (e-mail: \href{mailto:michael.neri@tuni.fi}{michael.neri@tuni.fi})}. 

\thanks{F. Battisti is with the Department of Information Engineering, University of Padova, Via Gradenigo 6/b, 35131, Padova, Italy. (e-mail: \href{mailto:federica.battisti@unipd.it}{federica.battisti@unipd.it}).}
\thanks{The work of F. Battisti was carried out within the “HEAT – Hybrid Extended
reAliTy” Project GA 101135637 funded by the EU Horizon Europe
Framework Programme (HORIZON).}
}

\markboth{IEEE Transactions on Broadcasting,~Vol.~X, No.~X, March~2025}%
{Neri \MakeLowercase{\textit{et al.}}: A Sample Article Using IEEEtran.cls for IEEE Journals}


\maketitle

\begin{abstract}
During the compression, transmission, and rendering of point clouds, various artifacts are introduced, affecting the quality perceived by the end user. However, evaluating the impact of these distortions on the overall quality is a challenging task. This study introduces PST-PCQA, a no-reference point cloud quality metric based on a low-complexity, learning-based framework. It evaluates point cloud quality by analyzing individual patches,  integrating local and global features to predict the Mean Opinion Score. 
In summary, the process involves extracting features from patches, combining them, and using correlation weights to predict the overall quality. This approach allows us to assess point cloud quality without relying on a reference point cloud, making it particularly useful in scenarios where reference data is unavailable.
Experimental tests on three state-of-the-art datasets show good prediction capabilities of PST-PCQA, through the analysis of different feature pooling strategies and its ability to generalize across different datasets. The ablation study confirms the benefits of evaluating quality on a patch-by-patch basis. Additionally, PST-PCQA's light-weight structure, with a small number of parameters to learn, makes it well-suited for real-time applications and devices with limited computational capacity. For reproducibility purposes, we made code, model, and pretrained weights available at \url{https://github.com/michaelneri/PST-PCQA}. 
\end{abstract}


\section{Introduction}

\IEEEPARstart{I}N recent years, thanks to the increasing capability of 3D acquisition systems, point clouds have emerged as one of the most popular formats for immersive media~\cite{Ak_TMM_2024}. Point clouds consist of a collection of points defined by geometric coordinates and optional attributes such as color and reflectivity. They provide the users with a more immersive experience than 2D content thanks to a realistic visualization and the possibility of interaction~\cite{Alexiou_2017_MMSP}.

Point clouds might undergo several distortions during acquisition, transmission, and display~\cite{Su_TIP_2023}. Acquisition distortions refer to errors and inaccuracies that occur during the capture of 3D data, typically from sensors like \ac{LiDAR} or structured light cameras. When transmitting these data over networks, compression is often necessary to reduce the file size, thus introducing artifacts like noise, resolution loss, or geometric inaccuracies~\cite{Liu_MNET_2021, Tu_TIM_2023}. During the display phase, hardware limitations or rendering algorithms may further affect the quality, potentially resulting in visual inconsistencies or inaccuracies~\cite{Tu_TIM_2023}. \Copy{distortion}{While compression artifacts~\cite{Yang_ToB_2023} are the most common distortions affecting the rendered point cloud, several other types of distortions can occur, which usually affect geometry and color consistency by introducing noise~\cite{Lian_ToB_2024}, degrading the overall visual quality of the content.}

\begin{figure}[t]
    \centering
    \includegraphics[width=1\linewidth]{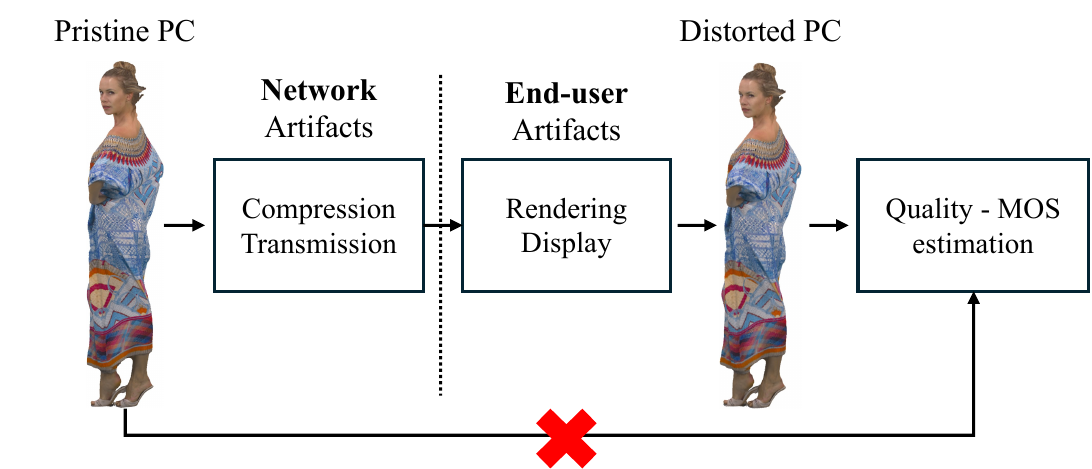}
    \caption{\Copy{caption}{Description of the no-reference point cloud quality assessment task. From acquisition to  rendering, the pristine point cloud is subject to several distortions that may impact the quality perceived by the user.}}
    \label{fig:task}
\end{figure}

Given that human observers are the primary users of point clouds in numerous applications, employing subjective quality assessment emerges as the most direct and dependable method for evaluating the quality of point clouds~\cite{Liu_2022_ACMTMCCA}. Despite its significance, subjective quality evaluation poses challenges due to its time-consuming nature and high cost. For the practical implementation of quality-focused point cloud systems, there is a strong demand for objective \ac{PCQA} models capable of accurately predicting subjective quality assessments~\cite{Liu_TCSVT_2021}. 

Objective quality estimators can be categorized into three classes: \ac{FR}, \ac{RR}, and \ac{NR}. \ac{FR} metrics assess quality by comparing the target against an unaltered original, requiring complete access to original data. \ac{RR} methods need only partial original data, e.g., compression parameters, thus balancing accuracy with data accessibility. \ac{NR} architectures, instead, evaluate quality without any reference to the original, offering flexibility in real scenarios but potentially at the cost of precision (Figure~\ref{fig:task}). Moreover, the availability of pristine information may be difficult at the end user device, especially in broadcasting and telecommunication scenarios, thus motivating the development of no-reference metrics for immersive multimedia~\cite{Lamichhane_ToB_2023}. However, currently the literature lacks no-reference methods for efficient estimation of the quality of distorted point clouds~\cite{Wang_TCSVT_2024, Liu_2022_ACMTMCCA}.
\Copy{examples}{
\begin{figure*}[ht]
    \centering
     \subfloat[V-PCC]{\includegraphics[width=0.2\linewidth]{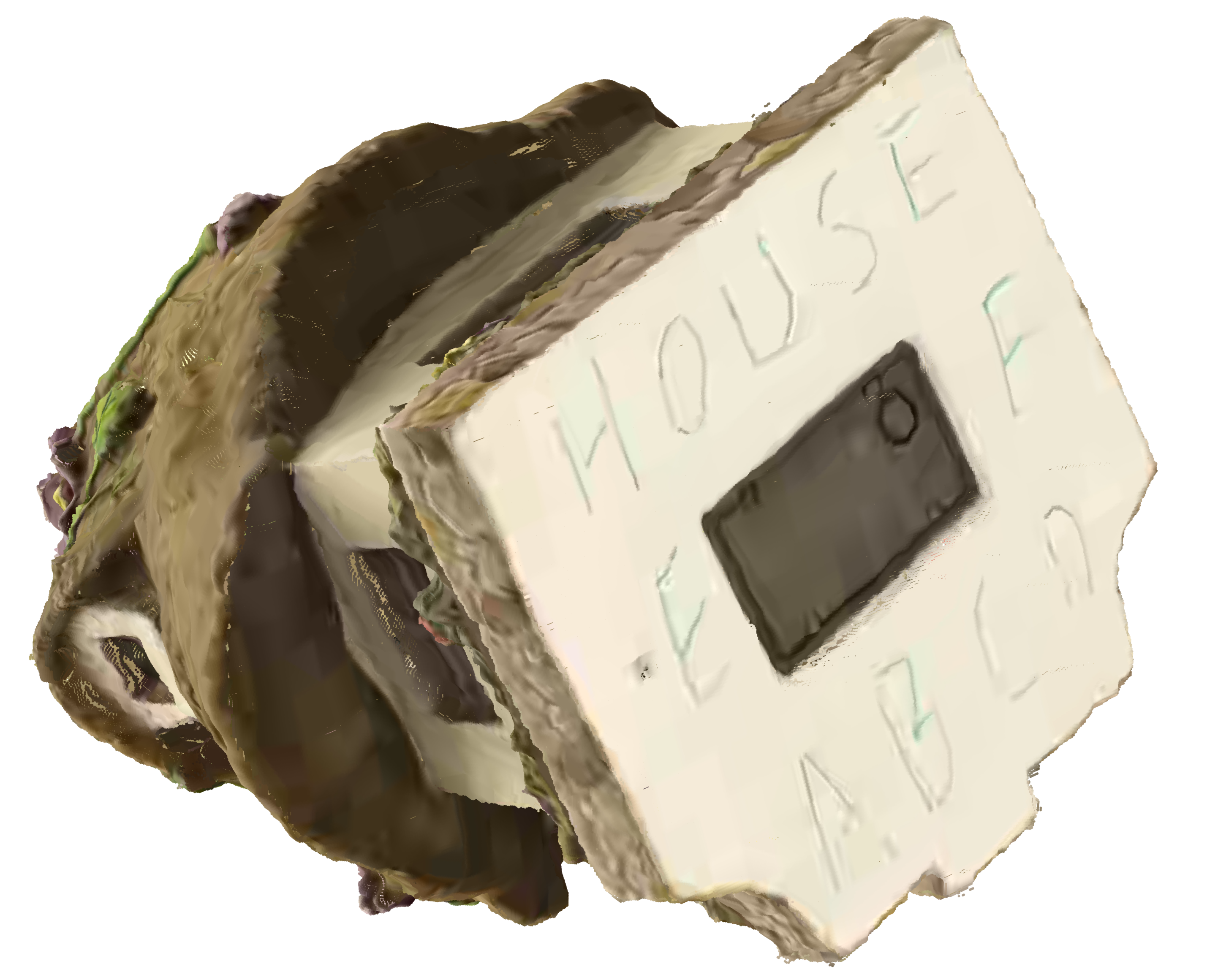}\label{fig:V-PCC}}
     \subfloat[Gaussian noise]
  {\includegraphics[width=0.2\linewidth]{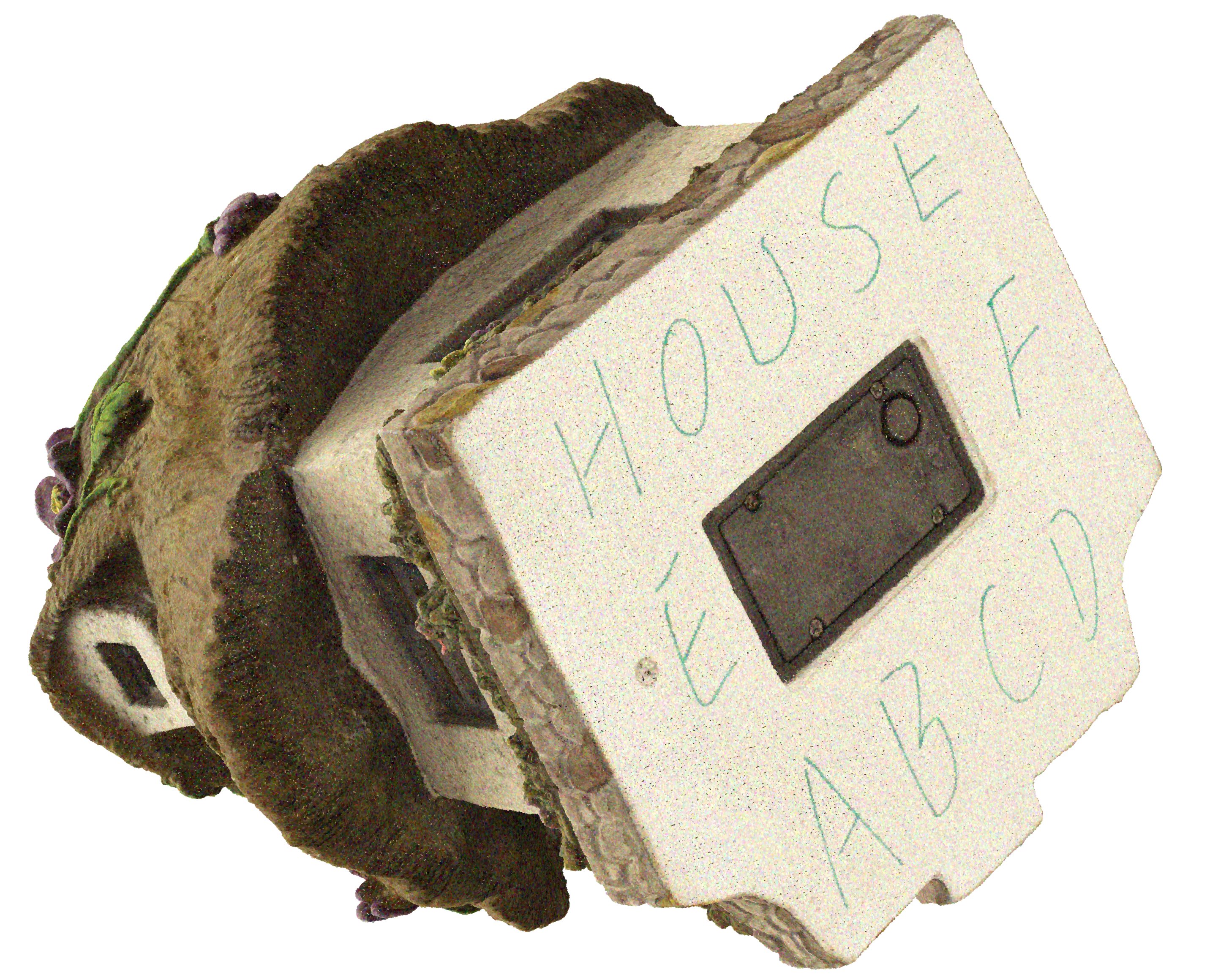}\label{fig:GN}}
     \subfloat[Downsampling]{\includegraphics[width=0.2\linewidth]{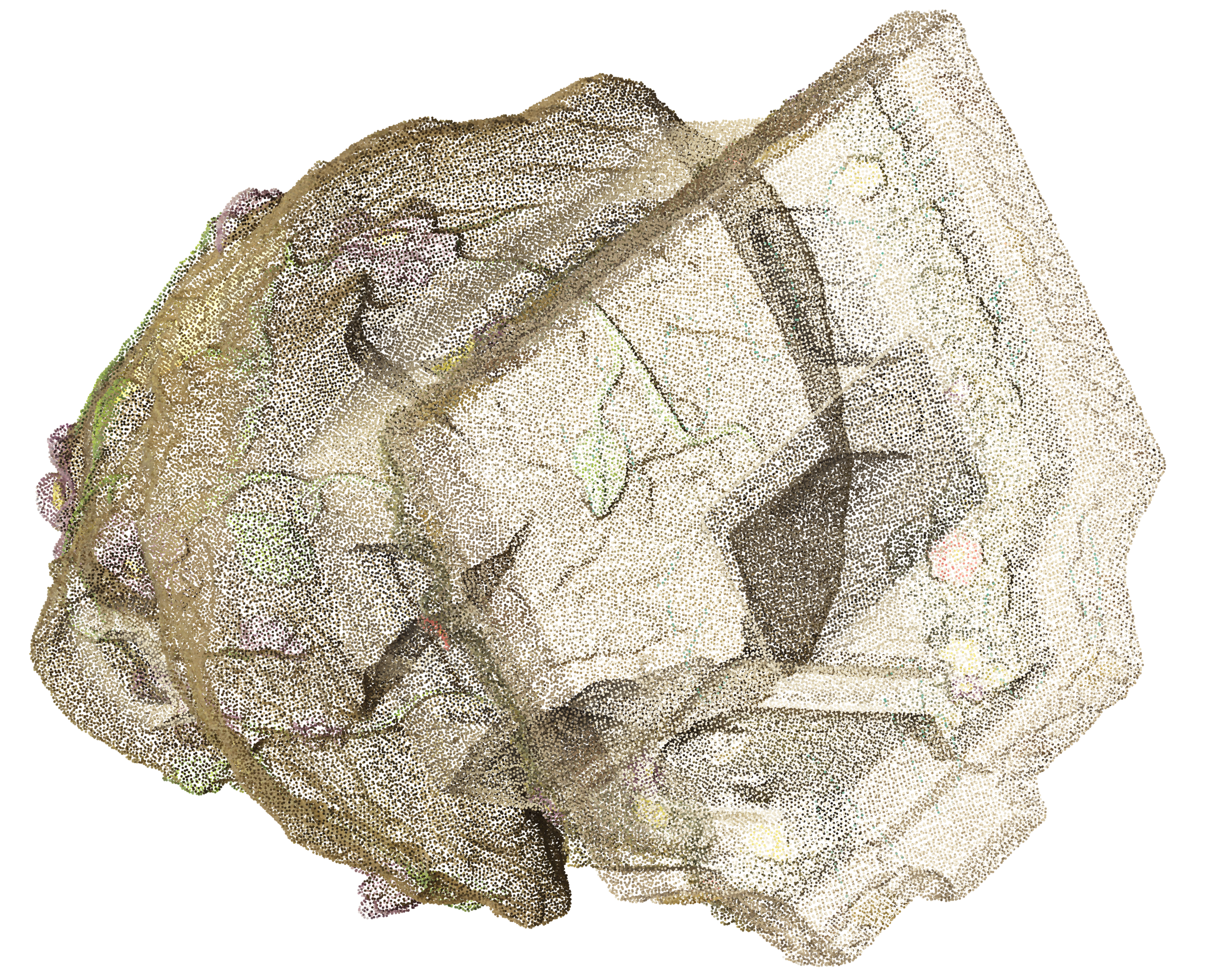}\label{fig:DW}}
     \subfloat[G-PCC]{\includegraphics[width=0.2\linewidth]{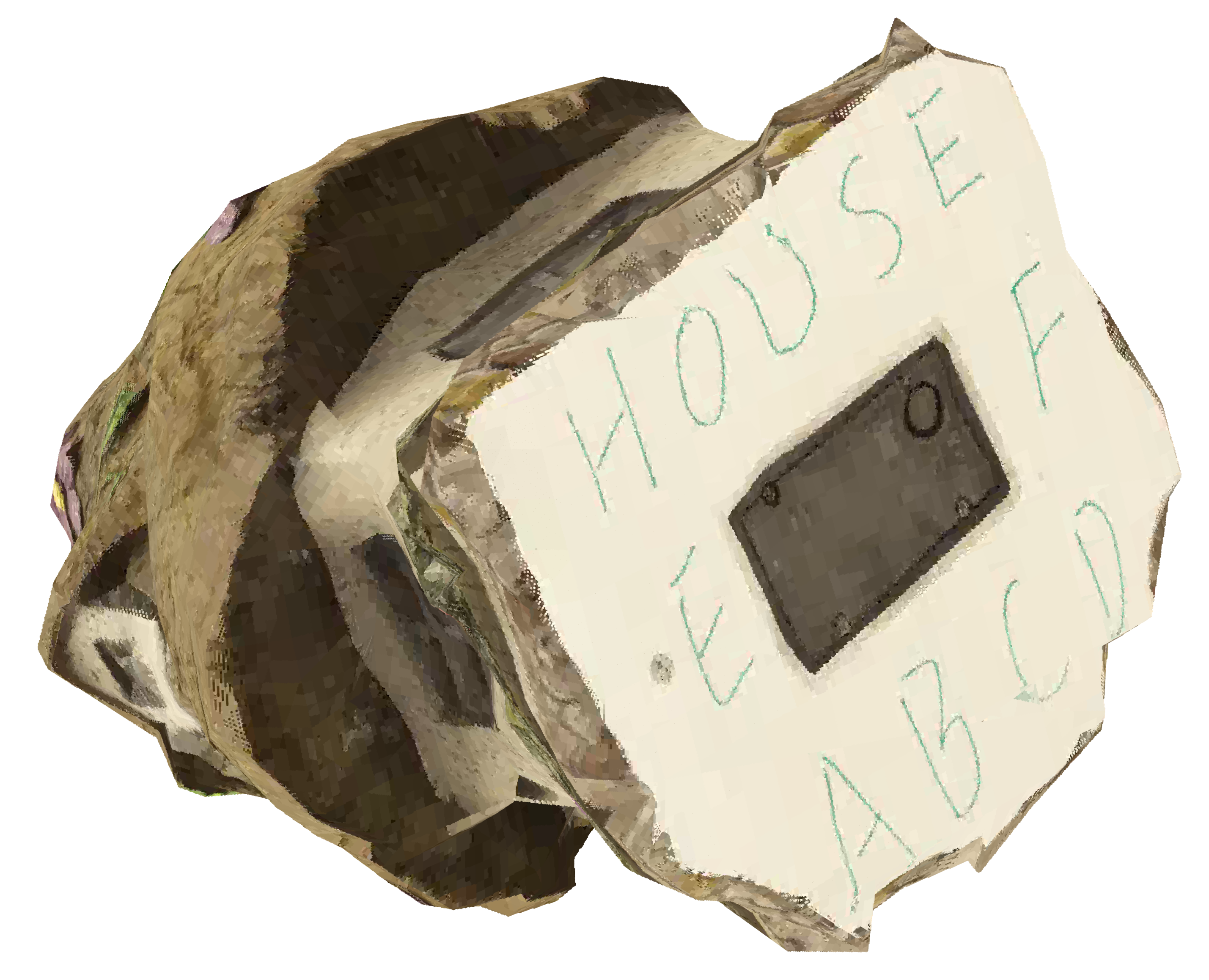}\label{fig:G-PCC}}
    \caption{Examples of compression techniques applied to the \textit{House} point cloud from WPC~\cite{Liu_TVCG_2023} dataset: (a) V-PCC with 'geometryQP'= $35$ and 'textureQP'= $45$; (b) Gaussian noise with standard deviation = $0$ for points' coordinates and $16$ for points' RGB values; (c) downsampling uniformly dividing the point cloud in $2^8$ segments; (d) G-PCC with trisoup, 'NodeSizeLog2'= $4$ and RAHT quantization step= $64$. }
    \label{fig:compressions}
\end{figure*} 
}


\Copy{soa}{State-of-the-art NR PCQA metrics present several challenges:
\begin{itemize}
    \item \textit{non-ML methods exhibit low correlation between predicted and ground truth quality scores~\cite{Mittal_SPL_2013, Mittal_TIP_2012};}
    \item \textit{methods exploiting specific features extracted from the point clouds (e.g., texture and structure) are mainly working in a global way on the entire point cloud~\cite{Liang_ToB_2024};}
    \item \textit{deep learning-based NR PCQA require a significant amount of computational resources and available datasets (also needed to reduce generalization issues)~\cite{Zhang_IJCAI_2023, Zhang_ICME_2023, Chai_TCSVT_2024}}. 
\end{itemize}}

\Copy{sparse}{In this work, we introduce PST-PCQA, a low-complexity learning-based NR PCQA for non-sparse point clouds that outperforms state-of-the-art architectures. In more detail, PST-PCQA splits point cloud into patches from which texture and structure features are extracted.} Those features are then combined to predict the overall quality. This approach is lightweight, i.e., the number of learnable parameters ($1.8$M) is the lowest in the state-of-the-art, with a total decrease of $93$\% with respect to the most efficient approach. This characteristic is crucial in devices where the computational load is limited and in applications where system response time should be real-time. 

\begin{figure*}[ht!]
    \centering
    \includegraphics[width=0.8\linewidth]{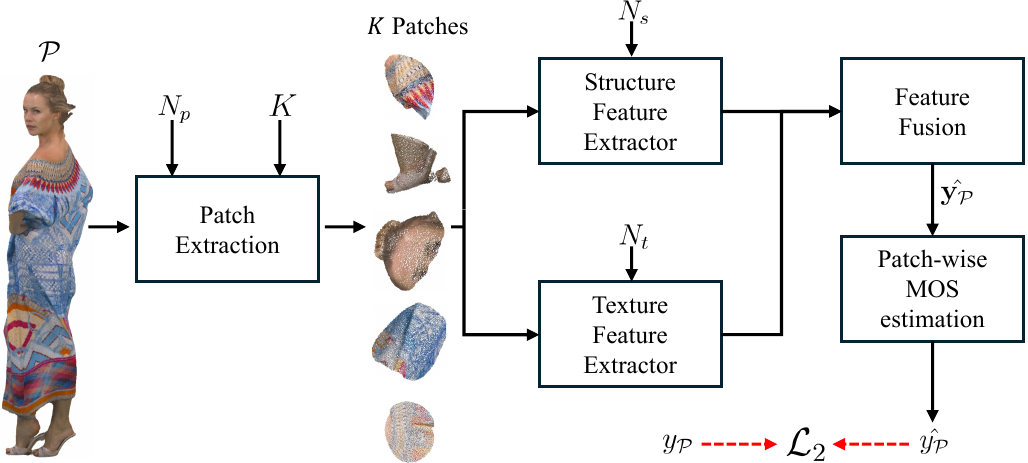}
    \caption{Description of the proposed approach. 
    }
    \label{fig:approach}
\end{figure*}

The main contributions of this paper are as follows:
\begin{itemize}
    \item the definition of a new lightweight \ac{NR} \ac{PCQA}, namely \ac{PST}-\ac{PCQA}, that exploits texture and structure features of the point cloud;
    \item a patch-wise quality estimation strategy. This approach allows the adoption of learned weights per patch and to improve explainability;
    \item an extensive analysis on state-of-the-art datasets for \ac{NR} \ac{PCQA} to demonstrate the effectiveness of the approach. Comparisons with other methods in the literature are carried out.
\end{itemize}
The remainder of this paper is structured as follows: Section~\ref{sec:related} details the relevant works in the literature. Section~\ref{sec:method} illustrates the proposed approach from the extraction of features to the final prediction of the quality. Section~\ref{sec:exp} reports the performance of our \ac{NR} \ac{PCQA} metric on $3$ state-of-the-art datasets, providing insights on the design rationale of the approach. Finally, Section~\ref{sec:conclusion} draws the conclusions with possible future directions of the work.

\section{Related Works}\label{sec:related}
In this section, state-of-the-art metrics for the quality assessment of point clouds are presented. First, \ac{FR} and \ac{RR} approaches are introduced. Then, an in-depth description of existing \ac{NR} metrics is provided. 

\subsection{Full- and reduced-reference metrics}
In~\cite{Meynet_QoMEX_2020}, the first attempt to assess the quality of colored point clouds was proposed. In more detail, the model PCQM inspects both geometry-based (e.g., mean curvature) and color-based features (e.g., lightness, chroma, and hue) to predict the \ac{MOS} of a distorted point cloud. In this direction, in~\cite{Lu_SPL_2022}, the authors proposed TDESM, a \ac{FR} metric which employs 3D Difference of Gaussian filters on both reference and distorted point clouds to extract similarity features from edge information. The authors in~\cite{Liu_TVCG_2023} proposed a \ac{FR} metric that exploits 2D projections of the point cloud, which are then analyzed using IW-SSIM for predicting the overall quality. Instead, Zhang \textit{et al.}~\cite{Zhang_TVCG_2023} devised TCDM, a space-aware vector autoregressive model that defines the quality of a distorted point as the difficulty of transforming it into its corresponding reference.

With the advent of deep learning, the research community has started investigating the use of neural networks for quality assessment. An example of using learning-based methods for assessing the quality of point clouds was designed in~\cite{Chetouani_MMSP_2021}. Specifically, a \ac{VGG}-like \ac{CNN} randomly extracts patches from both pristine and distorted point clouds to analyze both structure and color characteristics. \ac{MOS} was then predicted by means of \acp{MLP}. In~\cite{Yang_TPAMI_2023}, the authors devised \ac{MPED} in which the differences between pristine and distorted point clouds were measured via a multiscale potential energy approach, inspired by classical physics.

To the best of our knowledge, two reduced reference metrics are available in the state-of-the-art. In~\cite{Liu_VCIP_2022} the authors exploited information of the artifact type (e.g., \ac{V-PCC} compression parameters) to predict the \ac{MOS} of the distorted point cloud. Moreover, in~\cite{Liu_TIP_2021} \ac{V-PCC} parameters are estimated from the original and the distorted point cloud to predict the \ac{MOS}.

As stated before, \ac{FR} and \ac{RR} metrics are effective but hardly applicable in real scenarios due to the unavailability of the pristine point cloud at the receiver. 

\subsection{No-reference metrics}
Similarly to \ac{FR} and \ac{RR} metrics, the research community had initially started investigating the quality of distorted point clouds by extracting hard-engineered features, i.e., characteristics coming from by domain knowledge and expertise. For instance, in~\cite{Hua_BMSB_2021}, the authors proposed BQE-CVP, a \ac{NR} metric that extracts features from the distorted point cloud such as geometric and color information. \ac{MOS} is then predicted using a \ac{RF}. Similarly, in~\cite{Zhang_TCSVT_2022} the distorted point cloud was projected into quality-related geometry and color feature domains in order to apply \ac{NSS} and entropy-based features. Finally, a \ac{SVR} model was devised to regress the quality of the input point cloud. In this direction, Liu \textit{et al.}~\cite{Liu_TMM_2023} analyzed the relationship between \ac{V-PCC} texture quantization parameters and perceptual coding distortion, providing the basis for the definition of a bitstream-layer \ac{NR} model. However, extracting hand-crafted and compression-based features yielded poor performance. Hence, learning-based methods employing neural networks have shown improved performances with respect to traditional methods thanks to their ability to automatically extract relevant features for predicting point clouds' quality.

The first relevant work using deep learning was proposed in~\cite{Liu_TCSVT_2021}, where 2D projections of the distorted point cloud were processed by several \acp{CNN}, whose features were concatenated to predict the overall quality. Similarly, in~\cite{Yang_2022_CVPR}, IT-PCQA was devised to predict the \ac{MOS} of distorted point clouds by inspecting multi-perspective images. Training of the \ac{DNN} was carried out as a domain adaptation problem, exploiting the subjective scores available for 2D natural images datasets in the state-of-the-art and transferring this knowledge to the \ac{NR} \ac{PCQA} task. 

Together with the release of a large-scale \ac{NR} \ac{PCQA} dataset, the authors in~\cite{Liu_2022_ACMTMCCA} proposed a 3D \ac{CNN} which exploited sparse convolutions directly on points, namely ResSCNN. This is the first approach that tackles the computational complexity problem of \ac{NR} metrics in this field, as ResSCNN encompassed only $1.2$M learnable parameters. However, similarly to prior works, its performance on well-known datasets were insufficient to be directly employed in real applications.

In~\cite{Zhang_ICME_2023} the authors proposed EEP-3DQA which employs lightweight Swin-Transformer~\cite{Liu_ICCV_2021} as the backbone for feature extraction to predict the quality of both point clouds and mesh models. Similarly to~\cite{Liu_TCSVT_2021, Yang_2022_CVPR}, projections of the distorted 3D model are extracted from six standard viewpoints and then analyzed by the \ac{DNN}. An example of using \ac{GCN} in \ac{PCQA} was devised in~\cite{Shan_TVCG_2023} which attentively analyzes the structural and textural perturbations within point clouds. Moreover, the approach involves a multi-task framework that predicts both distortion type and degree, increasing its sensitivity to several distortion types. 

In~\cite{Zhang_TMM_2023} the authors proposed to process static and dynamic views from a moving camera to have a more comprehensive assessment of point cloud quality. Specifically, VQA\_PC consists in rotating the camera around the point cloud, extracting spatial and temporal features using deep learning models, and combining them to predict the overall quality of the distorted point cloud. Following the same approach, Wang \textit{et al.}~\cite{Wang_TCSVT_2024} designed MOD-PCQA that exploits multiscale feature extraction to evaluate point cloud quality from various observational distances. The \ac{DNN} incorporates a three-branch network structure designed to extract features from different scales, enhancing the model's ability to capture and analyze the perceptual quality of point clouds.

A combination of learning-based and traditional features was proposed in~\cite{Liang_ToB_2024}. Specifically, the architecture named MFE-Net integrates an adaptive feature extraction (AFE) module for local hand-crafted feature extraction, a local quality acquisition (LQA) model for deep feature learning, and a global quality acquisition (GQA) layer that aggregates these assessments into the  global predicted \ac{MOS}. Another proposed \ac{NR} metric is Plain-PCQA~\cite{Chai_TCSVT_2024}, where spatial geometric properties and texture details are jointly analyzed. 

Recent works, rather than processing 2D projections, employed patches from the point cloud to extract features and regress the \ac{MOS}~\cite{Cheng_ARXIV_2023, Zhang_IJCAI_2023}. 

Based on these new approaches, we propose a low-complexity deep learning metric that outperforms existing state-of-the-art models in predicting the \ac{MOS} of non-sparse distorted point clouds. Specifically, \ac{PST}-\ac{PCQA} splits the point cloud into patches to separately extract structure and texture features, which are then integrated to estimate the overall quality. Moreover, thanks to its lightweight design, this model can be effectively employed in environments with limited resources, differently from other learning-based approaches employing \ac{DNN}.

\begin{figure*}[ht]
    \centering
    \includegraphics[width=0.8\linewidth]{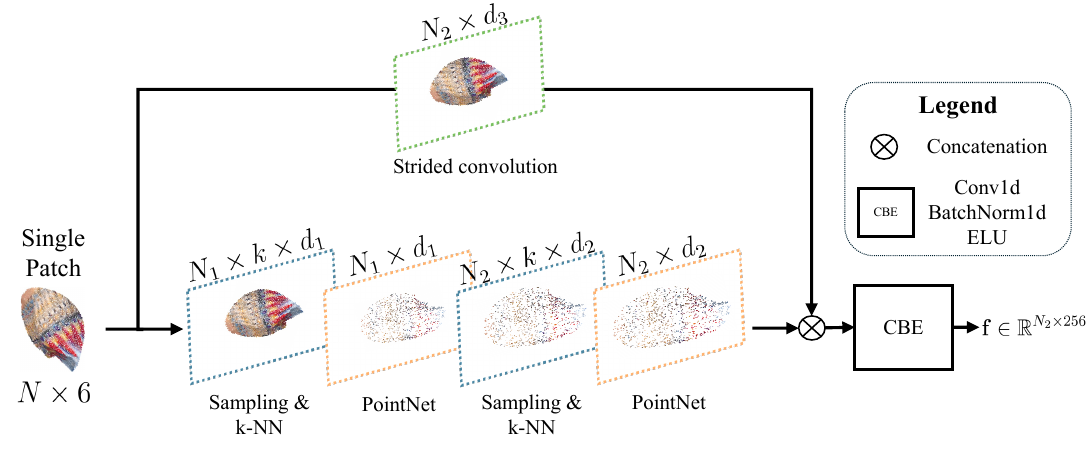}
    \caption{SFE and TFE neural architecture\modified{s}.}
    \label{fig:SFETFE}
\end{figure*}

\section{Proposed Approach}\label{sec:method}
The objective of this work is to estimate the quality of a generic point cloud as perceived by an average observer, the \ac{MOS}, without having information of its pristine version. Specifically, a point cloud is denoted as a set of $N$ points that represents the surface of a 3D object, i.e., $\mathcal{P} = \{ \mathbf{p_i}, i = 1, \ldots, N\}$. A single point is described as a vector containing its spatial coordinates and color information, $\mathbf{p_i} = [x_i, y_i, z_i, r_i, g_i, b_i]$. The set $\mathcal{P}$ can be denoted as a $N \times 6$ matrix, where $N$ is in the order of millions. Our approach aims at mapping the input point cloud and its quality $f:\mathbb{R}^{N \times 6} \rightarrow \mathbb{R}^+$ such as
\begin{equation}
    f(\mathcal{P}) = y_{\mathcal{P}},
\end{equation}
where $y_{\mathcal{P}} \in \mathbb{R}^+$ is the \ac{MOS} of the point cloud $\mathcal{P}$.

Figure~\ref{fig:approach} illustrates all the steps of the proposed architecture. Initially, a preprocessing step is applied to the distorted point cloud to obtain $K$ patches with $N_p$ points. Then, these portions of point cloud are fed to the \ac{SFE} and the \ac{TFE} modules to provide patch-wise features, analyzing both their structure and color patterns. Finally, a patch-wise and a global prediction of the quality of the distorted point cloud is provided as output. 


\subsection{Patch extraction}
\Copy{extraction}{Using point clouds patches can enhance computational efficiency and facilitate feature extraction as smaller data segments allow for more in-depth analysis and processing.  In fact, raw point clouds, especially those with high densities, may have millions of points, which can overwhelm memory and processing capabilities. Exploiting patches of point cloud can fasten the feature extraction process, whose characteristics can be combined by the MOS prediction module for both local and global analysis. Following this rationale, all points' coordinates $[x_i, y_i, z_i]$ of the point cloud $\mathcal{P}$ are first normalized in the range $1$ and $2001$, i.e., in a sphere with radius $1000$, for training stability and generalization purposes~\cite{Cheng_ARXIV_2023}. Then, FPS~\cite{Eldar_TIP_1997} and k-NN~\cite{Cover_TIT_1967} are employed to obtain $K$ centers and to sample $N_p$ points from the point cloud to compose the patches. Finally, all patches are concatenated along the channel dimension to compose the tensor with shape $K \times N_p \times 6$.}

\begin{figure*}[t]
    \centering
    \includegraphics[width=0.8\linewidth]{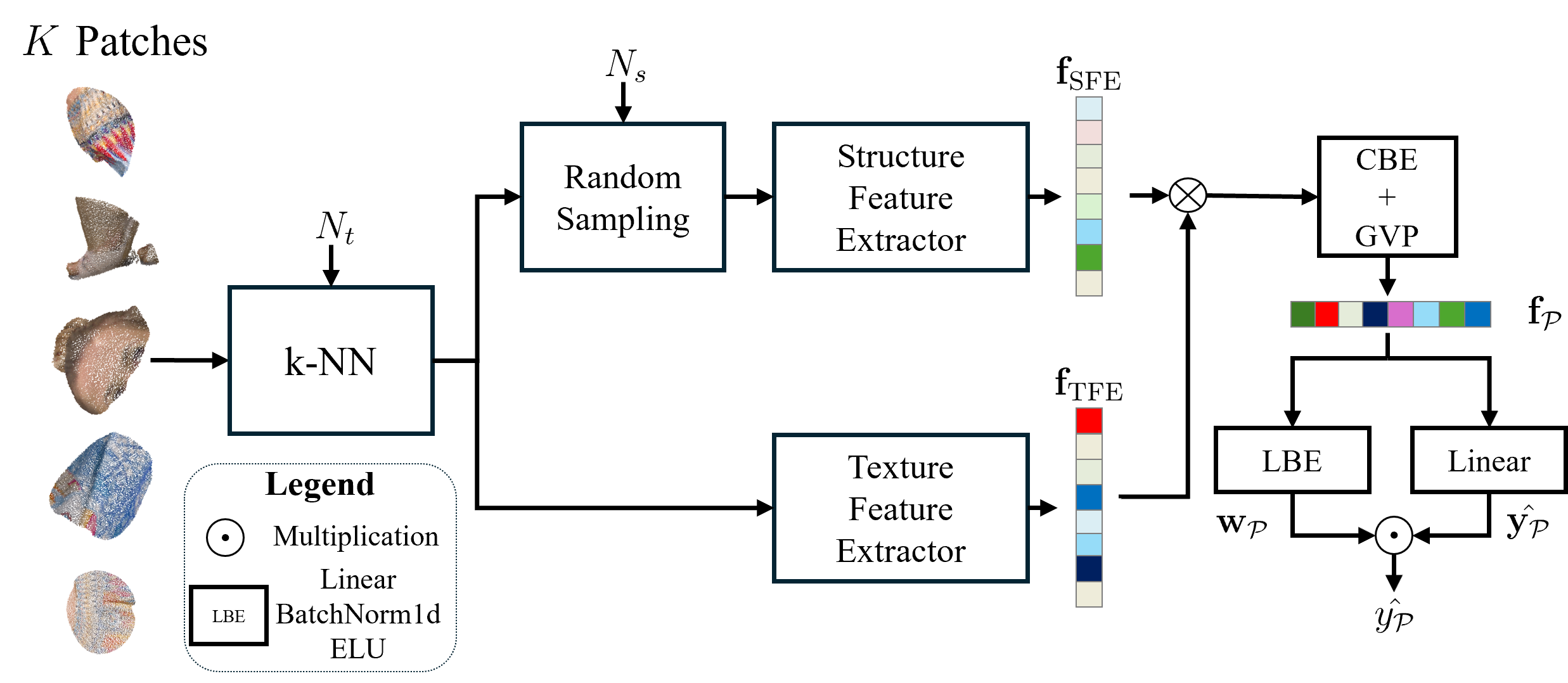}
    \caption{\ac{SFE} and \ac{TFE} common structure.}
    \label{fig:final_pred}
\end{figure*}

\subsection{Structure and texture feature extractors}
\ac{SFE} and \ac{TFE} neural networks adopt a layered feature extraction process, inspired by the hierarchical feature learning framework of PointNet++~\cite{Qi_NIPS_2017}. Specifically, they utilize the sampling, grouping, and PointNet (SGP) layers~\cite{Qi_NIPS_2017} to create an \textit{abstraction layer}, thus obtaining a transformed representation of the point cloud. This mechanism facilitates localized point cloud analysis that can be used to predict the \ac{MOS}. Our method differs from~\cite{Qi_NIPS_2017} in few key aspects:
\begin{itemize}
    \item we adopt grouped convolutions to reduce the total number of trainable parameters;
    \item we replace \ac{FPS} in the sampling phase with random sampling to enhance diversity and improve generalization capabilities;
    \item we employ the ELU~\cite{Clevert_ARXIV_2015} activation function instead of LeakyReLU. In fact, ELU leads to faster learning and to significantly better generalization performance than vanilla ReLUs and LeakyReLU on networks with more than 5 layers.
\end{itemize}

Figure~\ref{fig:SFETFE} depicts the structure of \ac{SFE} and \ac{TFE} that only differ in the number of input points. A patch with shape $N_t \times 6$ is fed to the \ac{TFE}, whereas its downsampled version $N_s \times 6$ is elaborated by the \ac{SFE}. By doing so, it is possible to simultaneously analyze patch's structural and texture features.

Each patch is analyzed by two sequential SGP layers. During the first pass, the patch is downsampled to $N_1$ points, grouped into $k$ centers by means of the \ac{k-NN}~\cite{Cover_TIT_1967} algorithm, and processed by the PointNet~\cite{Qi_NIPS_2017} layer, yielding the first transformed representation of the input patch, with shape $N_1 \times d_1$. The last \textit{abstraction layer} samples $N_2$ points, extracts $k$ centers, and projects each point to a $d_2$-dimensional space, resulting in a $N_2 \times d_2$ matrix.

In addition, the point cloud patch is analyzed by a learnable convolution with stride $s = \lfloor N/d_2 \rfloor$. The result of each branch is concatenated each other to obtain the patch's features $\mathbf{f} \in \mathbb{R}^{N_2 \times 256}$.

Finally, patch-wise features from both branches are arranged to compose texture and structure features of the input point cloud $\mathcal{P}$, namely $\mathbf{f_{\mathrm{SFE}}} \in \mathbb{R}^{K \times N_2 \times 256}$ and $\mathbf{f_{\mathrm{TFE}}} \in \mathbb{R}^{K \times N_2 \times 256}$ respectively.

\subsection{Patch-wise and global quality estimation}
To fuse the extracted characteristics from both \ac{SFE} and \ac{TFE} and yield a feature vector per patch $\mathbf{f}_{\mathcal{P}} \in \mathbb{R}^{K \times 512}$, we employ \ac{GVP} and a combination of Conv1d, batch normalization, and ELU (CBE) as follows:
\begin{equation}
    f_\mathcal{P} = \mathrm{CBE} (\mathrm{GVP}(\mathbf{f_{\mathrm{SFE}}} \otimes  \mathbf{f_{\mathrm{TFE}}})),
\end{equation}
where $\otimes$ denotes the concatenation function. Then, a LBE (linear-batchnorm1d-elu) and a linear layer are employed to predict patch-wise weights $\mathbf{w}_{\mathcal{P}} \in \mathbb{R}^{K}$ and scores $\hat{\mathbf{y}_{\mathcal{P}}} \in \mathbb{R}^{K}$ from  $\mathbf{f}_{\mathcal{P}}$
\begin{equation}
\begin{cases}
    \mathbf{w}_{\mathcal{P}} = \mathrm{LBE} (\mathbf{f}_{\mathcal{P}}) \\
    
    \hat{\mathbf{y}_{\mathcal{P}}} = \mathrm{Linear}(\mathbf{f}_{\mathcal{P}}).
    \end{cases}
\end{equation}
The predicted point cloud \ac{MOS} $\hat{y}_{\mathcal{P}}$ is then obtained by combining patch-wise scores with their weights
\begin{equation}
    \hat{y}_{\mathcal{P}} = \mathbb{E}_K[\mathbf{w}_{\mathcal{P}} \cdot \hat{\mathbf{y}_{\mathcal{P}}} ] 
\end{equation}
where $\mathbb{E}_K[\cdot]$ refers to the expected value across patches.

Figure~\ref{fig:final_pred} shows how the prediction of the point cloud quality $y_\mathcal{P}$ is estimated from its features $\mathbf{f}_{\mathrm{SFE}}$ and $\mathbf{f}_{\mathrm{TFE}}$. The model is trained by minimizing the \ac{MSE} of both patch-wise and global \ac{MOS} estimation with respect to the ground truth
\begin{equation}\label{eq:loss}
    \mathcal{L}(\hat{\mathbf{y}_{\mathcal{P}}}, \hat{y}_{\mathcal{P}}, y_{\mathcal{P}}) = \alpha \mathcal{L}_2 (\hat{\mathbf{y}_{\mathcal{P}}}, y_{\mathcal{P}}) + \beta \mathcal{L}_2 (\hat{y}_{\mathcal{P}}, y_{\mathcal{P}})
\end{equation}
where $\alpha \in \mathbb{R}^+$ and $\beta \in \mathbb{R}^+$ are two scalars for balancing the patch-wise and global \ac{MOS} estimation errors, respectively.

\section{Experimental Results}\label{sec:exp}

\begin{figure*}[ht!]
\centering
\begin{tabular}{cc}
  \includegraphics[width=.32\textwidth]{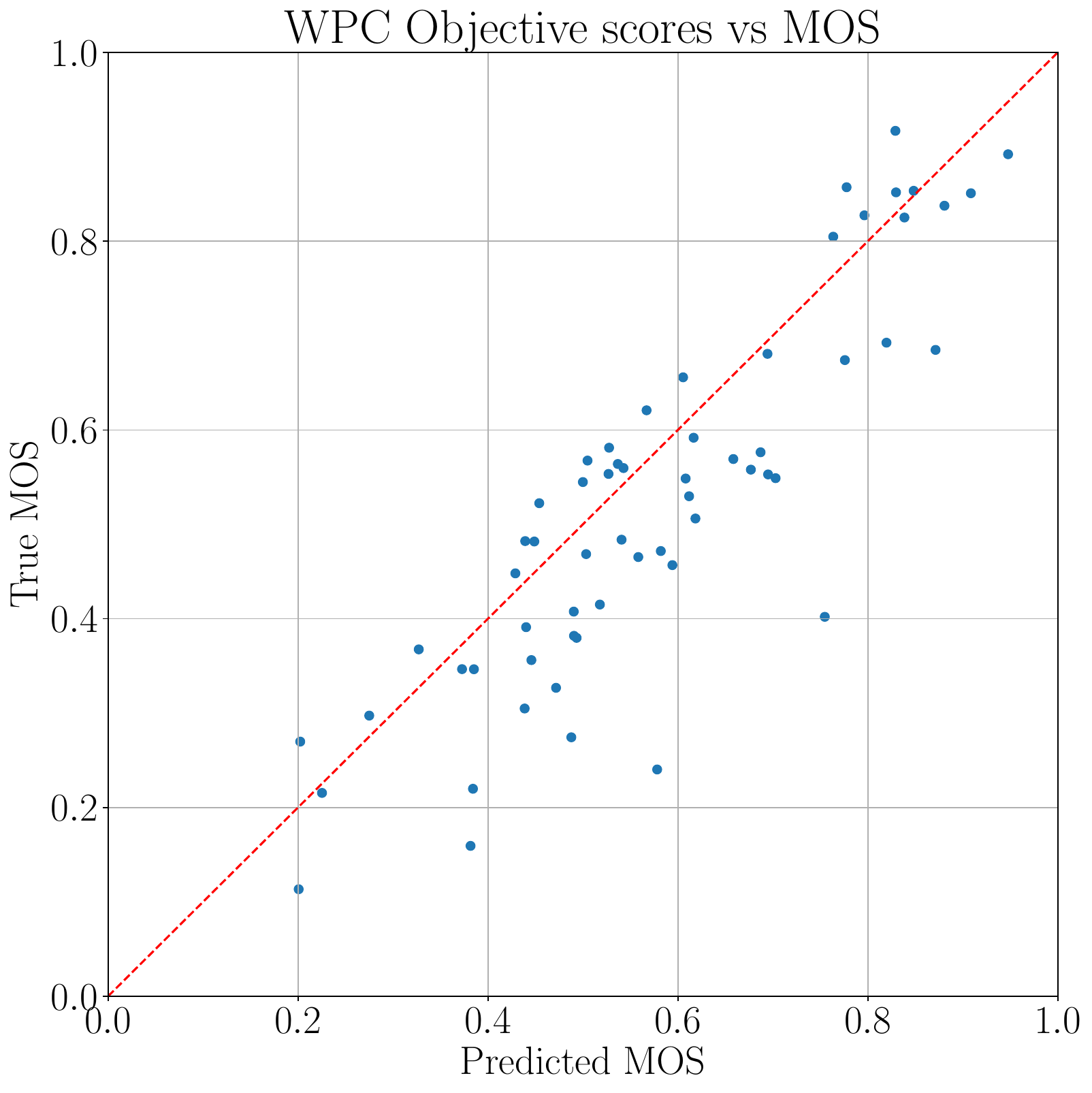}
  \includegraphics[width=.32\textwidth]{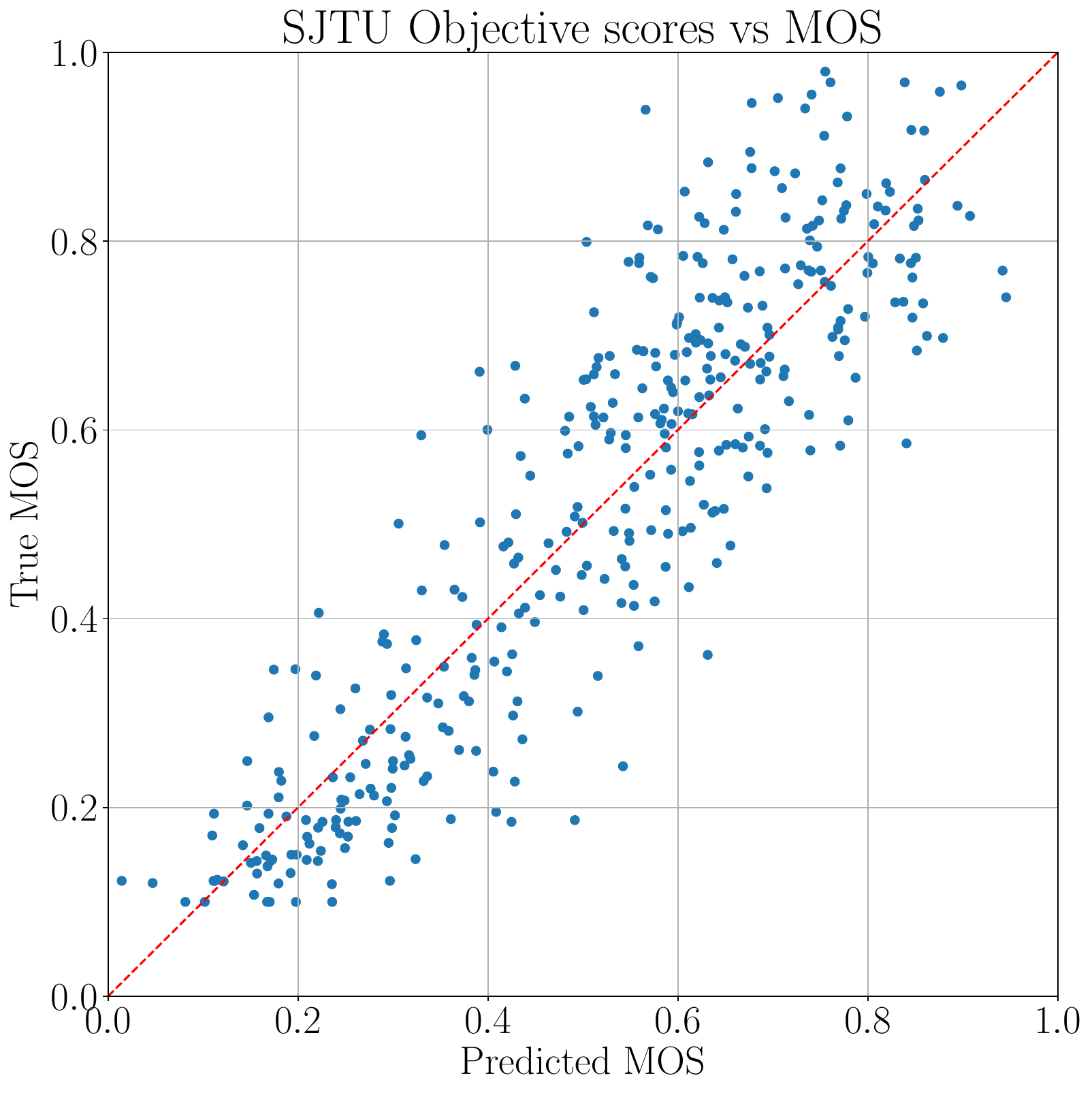}
  \includegraphics[width=.32\textwidth]{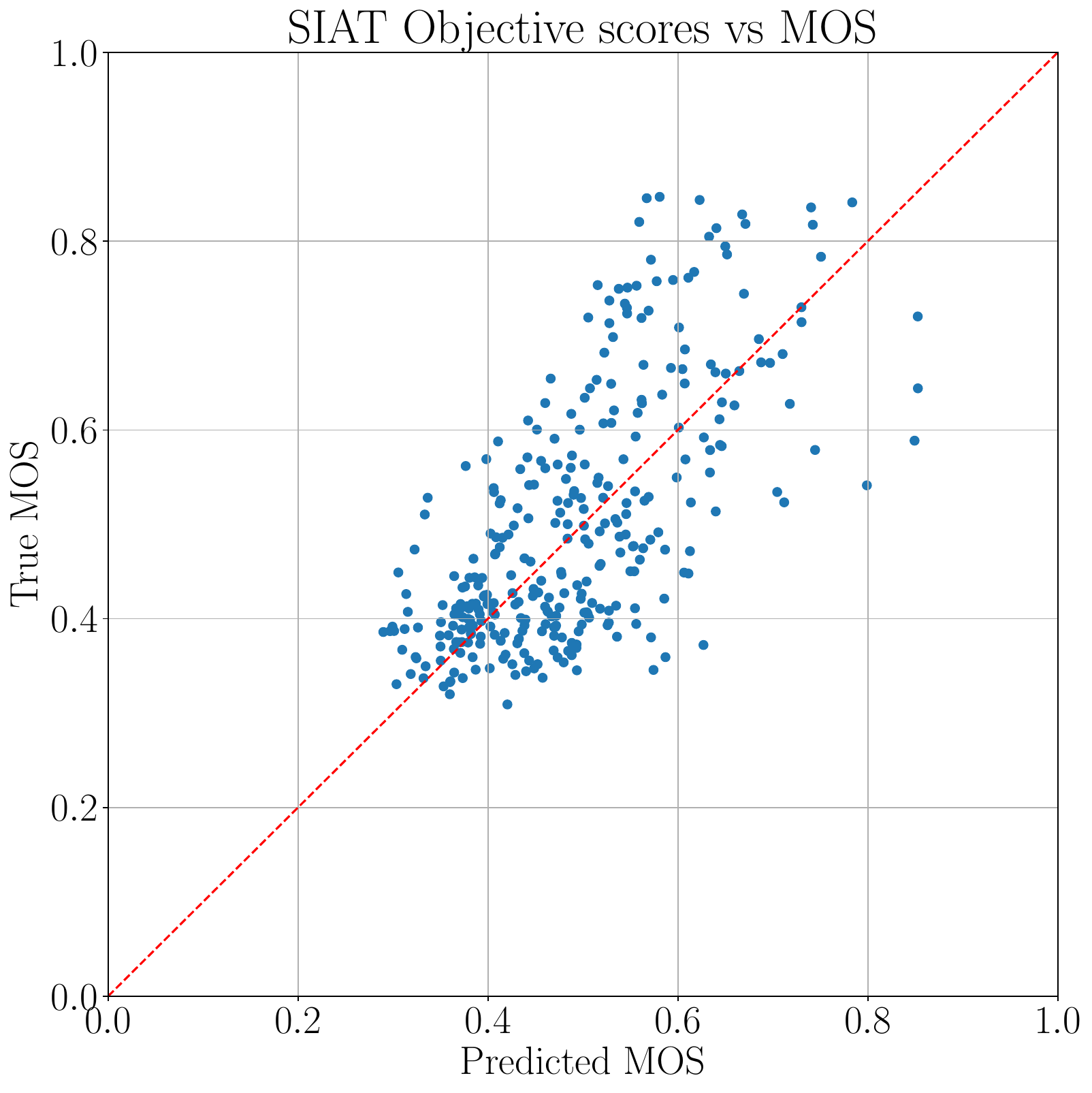}
    \end{tabular}
    \caption{Scatter plots between normalized predicted and ground truth MOS for WPC~\cite{Liu_TVCG_2023}, SJTU-PCQA~\cite{Yang_TMM_2021}, and SIAT-PCQD~\cite{Wu_TCSVT_2021} datasets, respectively. Identity line is plotted in red for comparison with ideal quality estimator.} 
    \label{fig:pred_mos}
\end{figure*}

\subsection{Datasets}
To assess the performance of PST-PCQA with respect to architectures in the literature, three state-of-the-art \ac{NR} \ac{PCQA} datasets are analyzed.

\textbf{WPC~\cite{Liu_TVCG_2023}.} It includes $20$ original reference point clouds, each subject to five distortions: Gaussian noise, downsampling, and three point cloud compression coding techniques proposed by MPEG (\ac{G-PCC} \textit{octave}, \ac{G-PCC} \textit{trisoup}, and \ac{V-PCC})~\cite{Schwarz_JETCAS_2019}. These distortions present a wide range of geometric and textural variations, offering substantial examples for learning. For every original reference point cloud, $37$ distorted versions are created, leading to a total of $740$ distorted point clouds (calculated as $37$ distortions multiplied by $20$ original samples) within the WPC database, all derived from $20$ original reference point clouds. \Copy{WPC}{WPC contains inanimate everyday objects (e.g., office supplies) with diverse geometric and textural complexity.}

\textbf{SIAT-PCQD~\cite{Wu_TCSVT_2021}.} The SIAT-PCQD database comprises $20$ reference point clouds, which undergo several preprocessing steps like subsampling, rotation, and scaling to achieve $10$-bit geometric precision. Each point cloud is then subject to distortions using different geometry parameters (ranging from $20$ to $32$ in $4$ increments) and texture parameters (ranging from $27$ to $42$ in $5$ increments) through the V-PCC~\cite{Schwarz_JETCAS_2019} coding method, resulting in 17 distinct distorted versions per reference point cloud. Consequently, the database encompasses a total of $340$ distorted point clouds. \Copy{SIAT-PCQD}{SIAT-PCQD contains both human figures and objects. The human category consists of six full-body figures and four upper-body figures, while objects include ten different instances (e.g.,  building).}

\textbf{SJTU-PCQA~\cite{Yang_TMM_2021}.} It comprises $10$ publicly accessible reference point clouds, each subject to $7$ types of synthetic distortions with $6$ intensity levels. These distortions include octree-based compression, color noise, geometry Gaussian noise, downscaling, and combinations thereof. Consequently, the SJTU-PCQA database features a total of $378$ distorted point clouds, obtained from $9$ samples multiplied by $7$ distortions and then by $6$ levels. This dataset has been included to analyze the performance of our approach on a dataset with few samples, thus evaluating its convergence stability. \Copy{SJTU-PCQA}{SJTU-PCQA includes six human models and four inanimate objects.}

All the distorted versions of the same point clouds are either in the training or in the testing dataset to avoid data leakage. Pooling selection and ablation studies are carried out on WPC~\cite{Liu_TVCG_2023} since, according to the literature, it is the most difficult \ac{NR} \ac{PCQA} dataset for data-driven approaches. In fact, WPC database incorporates more intricate distortions and exploits more levels of degradation, modeling real use cases. 

\subsection{Metrics}
The criteria for evaluating the relationship between predicted scores and quality labels are \ac{SRCC}, \ac{KRCC}, \ac{PLCC}, and \ac{RMSE}. A high-performing model is indicated by \ac{SRCC}, \ac{KRCC}, and \ac{PLCC} values approaching $1$, and a \ac{RMSE} value near $0$.

\subsection{Implementation details}
In this work, for fair comparison with state-of-the-art approaches, we split the datasets as follows:
\begin{itemize}
    \item \textbf{WPC:} We follow training and testing split as in~\cite{Liu_TCSVT_2021};
     \item \textbf{SIAT-PCQD:} Leave-one-out $20$ cross validation has been implemented;
    \item \textbf{SJTU-PCQA:} Leave-one-out $10$ cross validation has been adopted.
\end{itemize}

Following~\cite{Cheng_ARXIV_2023}, $K = 16$ patches with $N_p = 14900$ points are extracted from the distorted point clouds. Then, $N_s = 1024$ points are randomly sampled from the patch for analyzing its structure. Differently, $N_t = 8192$ points are selected by means of the \ac{k-NN} algorithm~\cite{Cover_TIT_1967}. In both \ac{TFE} and {SFE}, dimensionality of features are set to $d_1 = 128$ and $d_2 = 256$, with number of points $N_1 = 512$ and $N_2 = 256$, respectively. During SGP layers, the number of groups in the \ac{k-NN} algorithm is $k = 32$.  Overall, the number of trainable parameters of the learning-based metric is $1.8 \mathrm{M}$, highlighting the low-complexity of the approach. The model is trained for $400$ epochs with batches of size $4$. A cosine annealing learning rate is employed with initial learning rate $\eta_{\mathrm{max}} = 0.001$ with a maximum number of steps $T_{\mathrm{max}} = 400$. Both terms of the loss function in Eq. \eqref{eq:loss} equally contribute to the backpropagation algorithm ($\alpha = 1$ and $\beta = 1$). \Copy{environment}{The overall implementation and evaluation of PST-PCQA has been designed in Python 3.10 in a workstation with a CUDA-enabled graphic processing unit (NVIDIA RTX 4070).} Pytorch-Lightning and Weights\&Biases are utilized for training and logging, respectively. Further implementation details are available at \url{https://github.com/michaelneri/PST-PCQA}.


\subsection{Analysis on the type of pooling function}
Generally, the choice of pooling functions significantly affects the precision of the prediction in learning-based approaches~\cite{Zhou_TCSVT_2024}. To this aim, in Table~\ref{tab:studyPooling} an analysis on which type of feature extraction, both first- (e.g., \ac{GMP} and \ac{GAP}) and second-order (e.g., \ac{GVP}) statistical moments, works better on WPC~\cite{Liu_TVCG_2023} is presented. When combined, \ac{GAP} + \ac{GVP} and \ac{GMP} + \ac{GVP} show improved performance over \ac{GAP} and \ac{GMP} alone in terms of \ac{SRCC} and \ac{KRCC}. However, it is worth highlighting the effectiveness of \ac{GVP} for this task, with the highest correlation to MOS in terms of \ac{PLCC}, \ac{SRCC}, and \ac{KRCC}. This suggests that combining pooling methods can capture a broader range of features that may correlate with human perception, but the combination might not always lead to improvement. 
\begin{table}[ht!]
\caption{Performance of different feature pooling on the WPC dataset.}
\label{tab:studyPooling}
    \centering
        \adjustbox{max width=0.49\textwidth}{%
    \begin{tabular}{c|cccc}
    \hline \hline
        Pooling & \ac{PLCC} $\uparrow$ & \ac{SRCC} $\uparrow$ & \ac{KRCC} $\uparrow$ & \ac{RMSE} $\downarrow$  \\
    \hline
     \ac{GMP} & $0.8501$ & $0.8295$ & $0.6506$ & $10.4842$ \\
     \ac{GAP} & $0.8657$ & $0.8511$ & $0.6703$ & $\mathbf{10.2192}$ \\
     \ac{GAP} + \ac{GVP} & $0.8462$ & $0.8133$ & $0.6298$ & $10.4975$ \\
     \ac{GMP} + \ac{GVP} & $0.8712$ & $0.8463$ & $0.6696$ & $10.2588$ \\
     \ac{GVP} & $\mathbf{0.8821}$ & $\mathbf{0.8624}$ & $\mathbf{0.6854}$ & $10.5769$ \\
     \hline \hline
    \end{tabular}
}
\end{table}

\subsection{Analysis on the number of patches $K$}
\Copy{K_number}{Although PST-PCQA feature extraction architecture does not change with respect to the number of patches $K$, the patch-wise \ac{MOS} estimation module includes batch normalization across patches. Hence, to evaluate the effect of $K$, we provide the performance of PST-PCQA with respect to diverse values of $K = \{2, 4, 8, 16, 32\}$ in Table~\ref{tab:patches_study}. From the results it is worth noting that having few patches ($K = \{2, 4\}$) impacts the performance of PST-PCQA. Comparable results are achieved with $K = \{8, 16\}$ patches whereas having more than $K = 16$ yields an inefficient model both in terms of correlation between true and predicted \ac{MOS} and of computational complexity}. 

\Copy{table_k}{\begin{table}[ht!]
\caption{Performance of different number of patches $K$ on the WPC dataset.}
\label{tab:patches_study}
    \centering
    \begin{tabular}{c|c|c|c|c}
    \hline \hline
         $K$ & \ac{PLCC} $\uparrow$ & \ac{SRCC} $\uparrow$ & \ac{KRCC} $\uparrow$ & \ac{RMSE} $\downarrow$    \\ \hline
         $2$ & $0.8075$ & $0.7906$ & $0.6251$ & $13.1335$ \\
         $4$ & $0.8208$ & $0.8087$ & $0.6235$ & $12.5506$ \\
         $8$ & $0.8772$ & $\mathbf{0.8779}$ & $\mathbf{0.6965}$ & $10.7112$ \\
         $16$ & $\mathbf{0.8821}$ & $0.8624$ & $0.6854$ & $\mathbf{10.5769}$ \\
         $32$ & $0.8438$ & $0.8266$ & $0.6390$ & $11.7893$  \\
         \hline \hline
    \end{tabular}    
\end{table}}

\subsection{Results on all the datasets}
We compare the results of PST-PCQA with different types of state-of-the-art approaches:
\begin{itemize} 
    \item \ac{FR}: $\mathrm{p2p}_{\mathrm{MSE}}$~\cite{Mekuria_ISOIEC_2016}, $\mathrm{p2p}_{\mathrm{H}}$~\cite{Mekuria_ISOIEC_2016}, $\mathrm{p2plane}_{\mathrm{MSE}}$~\cite{Tian_2017_ICIP}, $\mathrm{p2plane}_{\mathrm{H}}$~\cite{Tian_2017_ICIP}, $\mathrm{PSNR}_{\mathrm{Y}}$~\cite{Mekuria_VCIP_2017}. IW-SSIM~\cite{Liu_TVCG_2023}, PCQM~\cite{Meynet_QoMEX_2020}, MPED~\cite{Yang_TPAMI_2023}, TCDM~\cite{Zhang_TVCG_2023}.
    \item \ac{RR}: $\mathrm{PCM}_{\mathrm{R}}$~\cite{Viola_SPL_2020} and Liu \textit{et al.}~\cite{Liu_TIP_2021}.
    \item \ac{NR}\footnote{COPP-Net~\cite{Cheng_ARXIV_2023} is not included due to incorrect training, validation, and testing splits, yielding incomparable results.}: BRISQUE~\cite{Mittal_TIP_2012}, NIQE~\cite{Mittal_SPL_2013}, ResCNN~\cite{Liu_2022_ACMTMCCA}, 3D-NSS~\cite{Zhang_TCSVT_2022}, PQA-Net~\cite{Liu_TCSVT_2021}, VQA-PC~\cite{Zhang_TMM_2023}, MM-PCQA~\cite{Zhang_IJCAI_2023}, EEP-3DQA\cite{Zhang_ICME_2023}, BEQ-CVP~\cite{Hua_BMSB_2021}, SGT-PCQA~\cite{Tu_TIM_2023}, and Plain-PCQA~\cite{Chai_TCSVT_2024}. 
\end{itemize}

\begin{table*}[ht!]
\caption{Experimental results on WPC and SJTU-PCQA. \textbf{bold} and  \underline{underline} notations have been used for highlighting the best and second performance, respectively. ($-$) means no data is available.}
\label{tab:resultsAllDatasets}
    \centering
        \adjustbox{max width=1\textwidth}{%
    \begin{tabular}{cc|cccc||cccc}
    \hline \hline
    & & \multicolumn{4}{c||}{WPC~\cite{Liu_TVCG_2023}} & \multicolumn{4}{c}{SJTU-PCQA~\cite{Yang_TMM_2021}} \\
        \multirow{2}*{} &  & \ac{PLCC} $\uparrow$ & \ac{SRCC} $\uparrow$ & \ac{KRCC} $\uparrow$ & \ac{RMSE} $\downarrow$ & \ac{PLCC} $\uparrow$ & \ac{SRCC} $\uparrow$ & \ac{KRCC} $\uparrow$ & \ac{RMSE} $\downarrow$  \\
    \hline
    \multirow{9}*{\ac{FR}} & $\mathrm{p2p}_{\mathrm{MSE}}$~\cite{Mekuria_ISOIEC_2016} & $0.4853$ & $0.4559$ & $0.3182$ & $19.8943$ & $0.8228$ & $0.7294$ & $0.5617$ & $1.3290$  \\
     & $\mathrm{p2p}_{\mathrm{H}}$~\cite{Mekuria_ISOIEC_2016} & $0.3972$ & $0.2786$ & $0.1944$ & $20.8993$ &  $0.8005$ & $0.7159$ & $0.5450$ & $1.3634$\\
     & $\mathrm{p2plane}_{\mathrm{MSE}}$~\cite{Tian_2017_ICIP} & $0.2440$ & $0.3282$ & $0.2250$ & $22.8226$ & $0.6697$ & $0.6278$ & $0.4825$ & $1.6961$\\
     & $\mathrm{p2plane}_{\mathrm{H}}$~\cite{Tian_2017_ICIP} & $0.3842$ & $0.2959$ & $0.2071$ & $21.0416$ & $0.7779$ & $0.6952$ & $0.5302$ & $1.4372$ \\

     & $\mathrm{PSNR}_{\mathrm{Y}}$~\cite{Mekuria_VCIP_2017} & $0.6166$ & $0.5823$ & $0.4164$ & $17.9001$ & $0.8124$ & $0.7871$ & $0.6116$ & $1.3224$ \\
     & IW-SSIM~\cite{Liu_TVCG_2023} & $0.8504$ & $0.8481$ & $-$ & $12.0600$ & $0.7949$ & $0.7833$ & $-$ & $1.4224$ \\
     & PCQM~\cite{Meynet_QoMEX_2020} & $0.6162$ & $0.5504$ & $0.4409$ & $17.9027$ & $0.8600$ & $0.8470$ & $-$ & $1.2370$ \\
     & MPED~\cite{Yang_TPAMI_2023} & $0.7000$ & $0.6780$ & $-$ & $16.3740$ & $0.8221$ & $0.7436$ & $0.5799$ & $1.2866$ \\
     & TCDM~\cite{Zhang_TVCG_2023} & $0.8070$ & $0.8040$ & $-$ & $13.5250$ & $0.9300$ & $0.9100$ & $-$ & $0.8910$ \\
    \hline \hline

    \multirow{1}*{\ac{RR}} & $\mathrm{PCM}_{\mathrm{R}}$~\cite{Viola_SPL_2020} & $0.3926$ & $0.3605$ & $0.2543$ & $20.9203$ & $0.6699$ & $0.5622$ & $0.4091$ & $1.7589$ \\
    \hline \hline
    \multirow{10}*{\ac{NR}} & BRISQUE~\cite{Mittal_TIP_2012} & $0.2614$ & $0.3155$ & $0.2088$ & $21.1730$ & $0.4214$ & $0.3975$  & $0.2966$ & $2.0930$ \\
     & NIQE~\cite{Mittal_SPL_2013} & $0.1136$ & $0.2225$ & $0.0953$ & $23.1410$ & $0.2420$ & $0.1379$ & $0.1009$ & $2.2620$\\
     & ResCNN~\cite{Liu_2022_ACMTMCCA} & $0.4531$ & $0.4362$ & $0.2987$ & $20.2591$ & $0.7821$ & $0.7911$ & $0.5224$ & $1.3651$  \\
     & 3D-NSS~\cite{Zhang_TCSVT_2022} & $0.6284$ & $0.6309$ & $0.4573$ & $18.1706$ & $0.7819$ & $0.7813$ & $0.6023$ & $1.7740$ \\
     & IT-PCQA~\cite{Yang_2022_CVPR} & $0.7950$ & $0.7800$ & $-$ & $-$ & $0.5800$ & $0.6300$ & $-$ & $-$\\
     & PQA-Net~\cite{Liu_TCSVT_2021} & $0.6671$ & $0.6368$ & $0.4684$ & $16.6758$ & $0.8586$  & $0.8372$ & $0.6304$ & $1.0719$  \\
     & VQA-PC~\cite{Zhang_TMM_2023} & $0.8001$ & $0.8012$ & $0.6237$ & $13.5570$  &  $0.8702$ & $0.8611$ & $0.6811$ & $1.1012$ \\
     & MM-PCQA~\cite{Zhang_IJCAI_2023} & $0.8556$ & $0.8414$ & $0.6513$ & $12.3506$ & $0.9226$ & $0.9102$ & $0.7838$ & $0.7716$ \\
     & EEP-3DQA~\cite{Zhang_ICME_2023} & $0.8296$ & $0.8264$ & $0.6422$ & $12.7451$ & $0.9363$ & $0.9095$ & $0.6811$ & $1.1010$\\
     & MOD-PCQA~\cite{Wang_TCSVT_2024} & $0.8733$ & \underline{$0.8752$} & $\mathbf{0.6952}$ & $11.0600$ & \underline{$0.9534$} & \underline{$0.9311$} & \underline{$0.7939$} & \underline{$0.7124$} \\
     & Plain-PCQA~\cite{Chai_TCSVT_2024} & \underline{$0.8783$} & $\mathbf{0.8793}$ & \underline{$0.6951$} & \underline{$10.8308$}  & $0.9302$ & $0.9133$ & $0.7603$ & $0.8607$ \\
     & \textbf{PST-PCQA} (ours) & $\mathbf{0.8821}$ & $0.8624$ & $0.6854$ & $\mathbf{10.5769}$ & $\mathbf{0.9593}$ & $\mathbf{0.9514}$ & $\mathbf{0.8049}$ & $\mathbf{0.6630}$\\
     \hline \hline
    \end{tabular}
}
\end{table*}

Table~\ref{tab:resultsAllDatasets} shows the comparison of performance between the proposed approach and the state-of-the-art on WPC~\cite{Liu_TVCG_2023} and SJTU-PCQA~\cite{Yang_TMM_2021} datasets. It is worth noticing the superiority of our approach on both datasets in mostly all the metrics with respect to \ac{NR} models, indicating a strong correlation with \ac{MOS}. \Copy{outperform}{Precisely, our approach outperforms other architectures in terms of \ac{PLCC} and \ac{RMSE} on the WPC dataset, whereas it surpasses the state-of-the-art on SJTU-PCQA in all the metrics.} In addition, PST-PCQA is also outperforming both \ac{RR} and \ac{FR} approaches, emphasizing its practical applicability in scenarios where the pristine point cloud is unavailable.  

Table~\ref{tab:resultsAllDatasets_SIAT} depicts the results of our method with respect to approaches in the literature on the SIAT-PCQD~\cite{Wu_TCSVT_2021} dataset, showing similar performance to SGT-PCQA~\cite{Tu_TIM_2023}. However, it is important to highlight the difficulty of learning-based approaches on this dataset due to the limited number of point cloud per distortions. In fact, best performance are obtained with hand-crafted features~\cite{Tu_TIM_2023, Hua_BMSB_2021} and machine learning regressors, such as \ac{RF}. \Copy{overfitting}{In our setup, $4$ folds were unsuccessful (\ac{PLCC} $\approx 50 \%$), whereas the others reached an average \ac{PLCC} performance of $90 \%$. This behavior is mainly caused by the model overfitting to the training set. To address this, designing data augmentation techniques or semi-supervised learning approaches in this field could further enhance performance.} 


To visually inspect the correlation between predicted and true \ac{MOS}, Figure~\ref{fig:pred_mos} displays the scatter plots across the three analyzed datasets.

\begin{table}[ht!]
\caption{Experimental results on SIAT-PCQD. ($-$) means no data is available. }
\label{tab:resultsAllDatasets_SIAT}
    \centering
        \adjustbox{max width=0.5\textwidth}{%
    \begin{tabular}{cc|cccc}
    \hline \hline
    &  &  \multicolumn{4}{c}{SIAT-PCQD~\cite{Wu_TCSVT_2021}}  \\
        \multirow{2}*{} &  & \ac{PLCC} $\uparrow$ & \ac{SRCC} $\uparrow$ & \ac{KRCC} $\uparrow$ & \ac{RMSE} $\downarrow$ \\
    \hline
    \multirow{9}*{\ac{FR}} & $\mathrm{p2p}_{\mathrm{MSE}}$~\cite{Mekuria_ISOIEC_2016} & $0.3136$ & $0.3963$ & $0.2761$ &  $0.1224$  \\
     & $\mathrm{p2p}_{\mathrm{H}}$~\cite{Mekuria_ISOIEC_2016} & $0.2980$ & $0.3791$ & $0.2620$ & $0.1231$ \\
     & $\mathrm{p2plane}_{\mathrm{MSE}}$~\cite{Tian_2017_ICIP} & $0.3498$ & $0.4125$ & $0.2947$ & $0.1208$\\
     & $\mathrm{p2plane}_{\mathrm{H}}$~\cite{Tian_2017_ICIP} & $0.3218$ & $0.3862$ & $0.2679$ & $0.1221$\\

     & $\mathrm{PSNR}_{\mathrm{Y}}$~\cite{Mekuria_VCIP_2017}  & $0.3443$ & $0.3481$ & $0.2318$ & $0.1211$  \\
     & IW-SSIM~\cite{Liu_TVCG_2023} & $0.8181$ & $0.6966$ & $0.5183$ & $0.0742$  \\
     & PCQM~\cite{Meynet_QoMEX_2020}  & $0.6539$ & $0.6666$ & $0.4825$ & $0.0994$ \\
    \hline \hline

    \multirow{2}*{\ac{RR}} & $\mathrm{PCM}_{\mathrm{R}}$~\cite{Viola_SPL_2020} & $0.3851$ & $0.3940$ & $-$ & $-$  \\
    & Liu \textit{et al.}~\cite{Liu_TIP_2021} & $0.9133$ & $0.9095$ & $-$ & $-$ \\
    \hline \hline
    \multirow{4}*{\ac{NR}} & 3D-NSS~\cite{Zhang_TCSVT_2022} & $0.5550$ & $0.5310$ & $-$ & $-$ \\
     & IT-PCQA~\cite{Yang_2022_CVPR} & $0.7870$ & $0.7920$ & $-$ & $-$ \\
     & SGT-PCQA~\cite{Tu_TIM_2023} & $\mathbf{0.8480}$ & $\mathbf{0.7950}$ & $-$ & $-$ \\
     & BEQ-CVP~\cite{Hua_BMSB_2021} & $0.7230$ & $0.6490$ & $-$ & $-$ \\
     & \textbf{PST-PCQA} (ours) & $0.8304$ & $0.7931$  & $\mathbf{0.5785}$ & $\mathbf{0.0183}$ \\
     
     \hline \hline
    \end{tabular}
}
\end{table}

\subsection{Cross-corpus generalization}
To assess the generalization capabilities of the proposed approach, we train on a source dataset, e.g., WPC, and test to a different dataset, e.g., SJTU-PCQA and viceversa. Table~\ref{tab:cross-corpus} depicts the results of PST-PCQA with respect to state-of-the-art approaches, demonstrating its ability to model \ac{HVS} of point clouds in out-domain scenarios. \Copy{cross}{In fact, PST-PCQA achieves the highest generalization performance in cross-dataset scenarios. For example, when trained on SJTU-PCQA and tested on WPC, it achieves an SRCC of 0.2737 and a PLCC of 0.3797, outperforming other methods such as VQA-PC (SRCC = 0.2733, PLCC = 0.3067). However, the generalization scores remain relatively low, reflecting the challenging nature of cross-corpus evaluation.}

\begin{table}[ht!]
\caption{Cross-corpus generalization analysis.}
\label{tab:cross-corpus}
    \centering
        \adjustbox{max width=0.5\textwidth}{%
    \begin{tabular}{c|cc|cc}
    \hline \hline
        & \multicolumn{2}{c|}{WPC $\rightarrow$ SJTU} & \multicolumn{2}{c}{SJTU $\rightarrow$ WPC} \\
        \multirow{2}*{} & \ac{SRCC} $\uparrow$ & \ac{PLCC} $\uparrow$ & \ac{SRCC} $\uparrow$ & \ac{PLCC} $\uparrow$  \\
    \hline
    3D-NSS~\cite{Zhang_TCSVT_2022} & $0.2117$ & $0.2034$ & $0.1214$ & $0.1313$ \\
    PQA-Net~\cite{Liu_TCSVT_2021} & $0.5411$ & $0.6102$ & $0.2211$ & $0.2334$  \\
    ResCNN~\cite{Liu_2022_ACMTMCCA} & $0.5012$ & $0.4954$ & $0.2301$ & $0.2293$ \\
    VQA-PC~\cite{Zhang_TMM_2023} & $0.5866$ & $0.6525$ & $0.2733$ & $0.3067$ \\
    \textbf{PST-PCQA} & $\mathbf{0.7413}$ & $\mathbf{0.7522}$ & $\mathbf{0.2737}$ & $\mathbf{0.3797}$  \\
    \hline \hline 
    \end{tabular}
}
\end{table}

\subsection{Ablation study}
To demonstrate the effectiveness of each component of the proposed approach, an ablation study has been carried out and the results are depicted in Table~\ref{tab:ablationstudyWPC}.  It is important to highlight the impact of patch-wise loss $\mathcal{L}_2 (\hat{\mathbf{y}_{\mathcal{P}}}, y_{\mathcal{P}})$, which acts as a regularizer for the model. Moreover, our approach without local weighting performs worse, with a decrease of the performance of $4.6\%$ in terms of \ac{PLCC}, validating our contribution. \Copy{ablation}{Finally, we evaluate the contribution of
each stream, namely \ac{TFE} and \ac{SFE}. The results show that the approach based on the fusion of the features obtained from TFE a SFE yields the best performance. However,
it is worth noting that when PST-PCQA includes only
one of the two streams, the obtained results are comparable. This
demonstrates that combining features with diverse point
densities improves the quality estimation.}

\Copy{ablation_WPC}{
\begin{table}[ht!]
\caption{Ablation study on the WPC dataset~\cite{Liu_TVCG_2023}.}
\label{tab:ablationstudyWPC}
    \centering
        \adjustbox{max width=0.5\textwidth}{%
    \begin{tabular}{c|cccc}
    \hline \hline
        \multirow{2}*{} & PLCC $\uparrow$ & SRCC $\uparrow$ & KRCC $\uparrow$ & RMSE $\downarrow$  \\
    \hline
    No $\mathcal{L}_2 (\hat{\mathbf{y}_{\mathcal{P}}}, y_{\mathcal{P}})$ & $0.7773$ & $0.6990$ & $0.5119$ & $13.6713$\\
    No LBE & $0.8358$ & $0.8587$ & $0.6667$ & $12.1790$ \\
    No TFE & $0.8642$ & $0.8639$ & $0.6783$ & $11.4367$ \\
    No SFE & $0.8751$ & $0.8637$ & $0.6836$ & $10.7429$ \\
    \hline
    \textbf{PST-PCQA} & $\mathbf{0.8821}$ & $\mathbf{0.8624}$ & $\mathbf{0.6854}$ & $\mathbf{10.5769}$ \\
    \hline \hline 
    \end{tabular}
}
\end{table}
}
\subsection{Complexity comparison}
Table~\ref{tab:complexity} compares the number of parameters of the proposed approach with the top-3 learning-based architectures in the literature. It is worth noting that PST-PCQA has the lowest number of learnable parameters, demonstrating its low-complexity nature. As stated in~\cite{Zhang_2021_CommunicationsACM}, a reduced number of trainable parameters can reduce the likelihood of overfitting, which is particularly beneficial for small datasets. Furthermore, models with fewer parameters demand less computational power and time during both the optimization and inference phase due to the decreased quantity of parameters. 

\Copy{3gpp}{According to 3GPP~\cite{3gpp.26.928} specification for Extended Reality (XR) applications, a system processing and broadcasting multimedia content is real-time if the delay is around or lower than $200$ ms. In our setup, PST-PCQA runs on a NVIDIA RTX 4070, which is a commercial-off-the-shelf GPU for gaming, with an average inference time of $70$ ms per point cloud.}

\begin{table}[ht!]
\caption{Complexity comparison with state-of-the-art.}
\label{tab:complexity}
    \centering
        \adjustbox{max width=0.48\textwidth}{%
    \begin{tabular}{c|cccc}
    \hline \hline
    Approach & MM-PCQA~\cite{Zhang_IJCAI_2023} & Plain-PCQA~\cite{Chai_TCSVT_2024} & EEP-3DQA~\cite{Zhang_ICME_2023} & \textbf{PST-PCQA}  \\
    \hline
    \# Params ($\mathrm{M}$) & $58.37$ & $28.50$ & $27.54$ & $\mathbf{1.80}$ \\
    \hline \hline
    \end{tabular}
}
\end{table}

\section{Conclusions}\label{sec:conclusion}
In this work we propose a novel low-complexity learning-based \ac{NR} \ac{PCQA}, namely \ac{PST}-\ac{PCQA}, which analyzes the input point cloud in patches. Our approach combines both local and global features to provide a prediction of the point cloud \ac{MOS}. Extensive experimental results on $3$ widely adopted datasets in the state-of-the-art show the effectiveness of \ac{PST}-\ac{PCQA}, providing design rationales on feature pooling, cross-corpus generalization capabilities. The ablation study demonstrates that a patch-wise analysis can enhance the performance of \ac{PST}-\ac{PCQA}. Furthemore, the reduced number of learnable parameters of \ac{PST}-\ac{PCQA} enables its use in real-time and computation-constrained hardware. \Copy{future}{A possible future investigation can concern the in-depth analysis of the impact of geometry- and texture-based distortions on the predicted \ac{MOS}.} Moreover, as a future work, adaptive 1D kernel convolutions, similarly to their 2D counterpart in~\cite{Zhou_TCSVT_2024}, i.e., changing kernel values with respect to the content, could be included to improve the generalization ability of \ac{PST}-\ac{PCQA}.

\bibliographystyle{IEEEbib}
\bibliography{bibliography}

\begin{IEEEbiography}[{\includegraphics[width=1in,height=1.25in,clip,keepaspectratio]{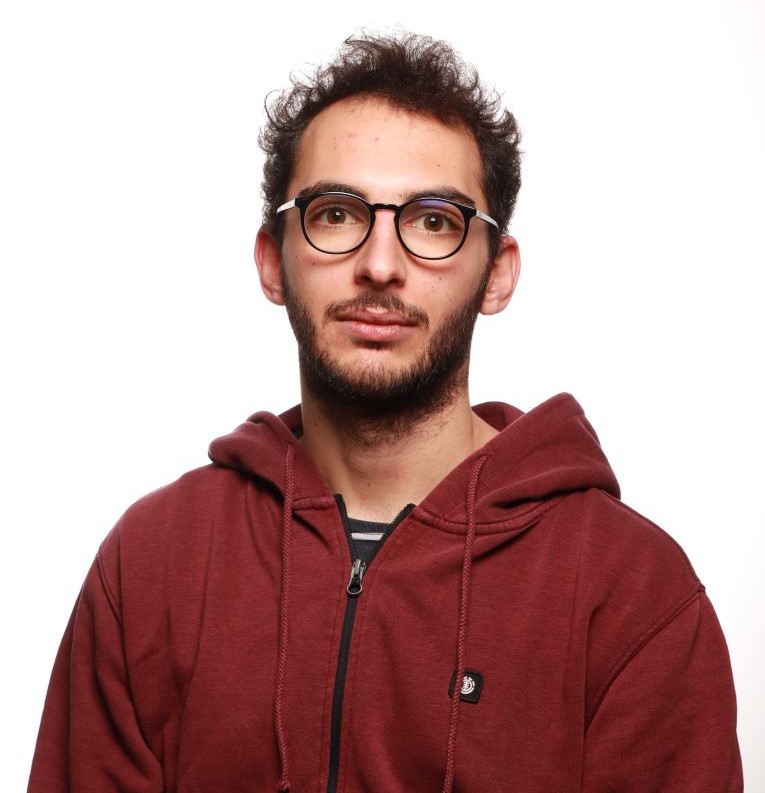}}]{Michael Neri} (Member, IEEE) obtained the Ph.D. in Applied Electronics (Roma Tre University) in 2025. He is now a researcher at Tampere University, Faculty of Information Technology and Communication Sciences, Finland. His main research interests are in the area of computer vision, deep learning, and audio processing. 
\end{IEEEbiography}

\begin{IEEEbiography}[{\includegraphics[width=1in,height=1.25in,clip,keepaspectratio]{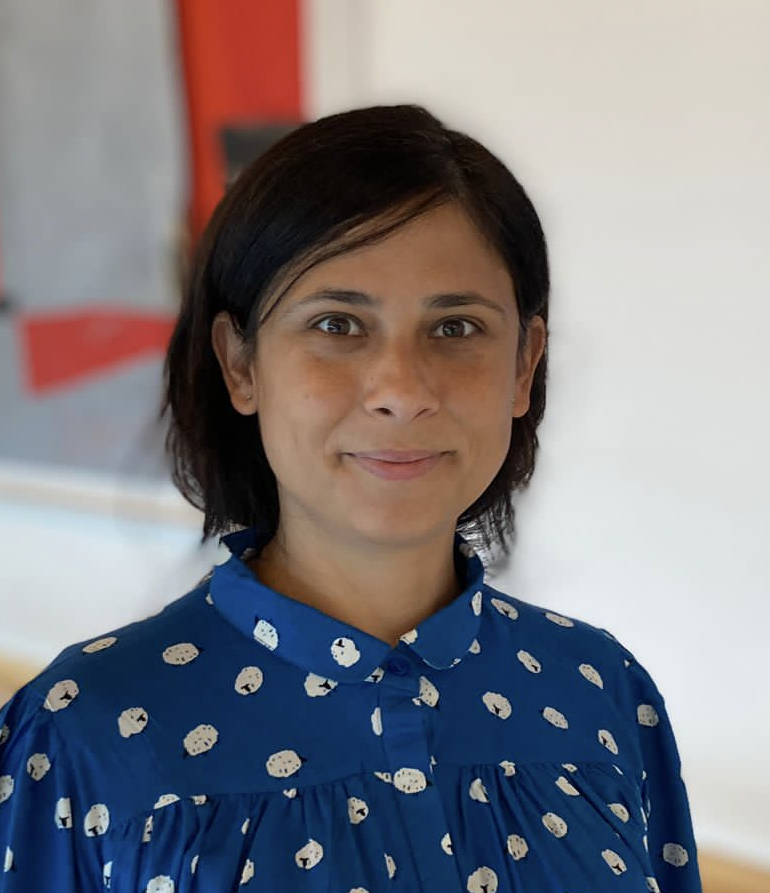}}]{Federica Battisti} (Senior Member, IEEE) is Associate Professor with the Department of Information Engineering, University of Padova. Her research interests include multimedia quality assessment and security. She is Editor in Chief for \textit{Signal Processing: Image Communication} (Elsevier) and Vice Chair of the EURASIP Technical Area Committee on Visual Information Processing. 
\end{IEEEbiography}

\end{document}